\documentclass[letterpaper]{article} 
\usepackage{aaai2026}  
\usepackage{times}  
\usepackage{helvet}  
\usepackage{courier}  
\usepackage[hyphens]{url}  
\usepackage{graphicx} 
\usepackage{natbib}  
\usepackage{caption} 
\usepackage[table]{xcolor} 
\usepackage{bbding}
\usepackage{algorithm}
\usepackage{graphicx}
\usepackage{amsmath}
\usepackage{amssymb}
\usepackage{mathtools}
\usepackage{amsthm}
\usepackage{multirow}
\usepackage{makecell}
\usepackage{algpseudocode}
\usepackage{booktabs} 
\usepackage{subfig}
\usepackage{tcolorbox}
\theoremstyle{plain}
\newtheorem{remark}{Remark}

\newtheorem{theorem}{Theorem}
\usepackage{svg}
\usepackage{microtype}      
\usepackage{xcolor}         
\definecolor{customgray}{gray}{0.9}
\usepackage{newfloat}
\usepackage{listings}
\DeclareCaptionStyle{ruled}{labelfont=normalfont,labelsep=colon,strut=off} 
\floatstyle{ruled}
\newfloat{listing}{tb}{lst}{}
\floatname{listing}{Listing}
\title{Calibrating and Rotating: A Unified Framework for Weight Conditioning in PEFT
}

\author {
     Da Chang\textsuperscript{\rm 1}\textsuperscript{\rm 2}\textsuperscript{\rm 4}, 
     Peng Xue\textsuperscript{\rm 1}\textsuperscript{\rm 2}\textsuperscript{\rm 4}, 
     Yu Li\textsuperscript{\rm 3}, 
     Yongxiang Liu\textsuperscript{\rm 1} \thanks{Correspondence author: \texttt{liuyx@pcl.ac.cn}}, 
     Pengxiang Xu\textsuperscript{\rm 1}, 
     Shixun Zhang\textsuperscript{\rm 1} \\
}
\affiliations {
    \textsuperscript{\rm 1}Pengcheng Laboratory,
    \textsuperscript{\rm 2}Shenzhen Institute of Advanced Technology, Chinese Academy of Sciences\\
    \textsuperscript{\rm 3}George Washington University,
    \textsuperscript{\rm 4}University of Chinese Academy of Sciences
}

\begin{document}
\maketitle
\begin{abstract}
Parameter-Efficient Fine-Tuning (PEFT) methods are crucial for adapting large pre-trained models. Among these, LoRA is considered a foundational approach. Building on this, the influential DoRA method enhances performance by decomposing weight updates into magnitude and direction. However, its underlying mechanism remains unclear, and it introduces significant computational overhead. In this work, we first identify that DoRA's success stems from its capacity to increase the singular value entropy of the weight update matrix, which promotes a more uniform update distribution akin to full fine-tuning. We then reformulate DoRA into a mathematically equivalent and more efficient matrix form, revealing it as a learnable weight conditioning method. Based on this insight, we propose a unified framework for designing advanced PEFT methods by exploring two orthogonal dimensions: the architectural placement and the transformation type of the conditioning matrix. Within this framework, we introduce two novel methods: (1) \textbf{Pre-Diag}, which applies a diagonal conditioning matrix before the LoRA update to efficiently calibrate the pre-trained weights, thereby enhancing performance while reducing training time; and (2) \textbf{S}kewed \textbf{O}rthogonal \textbf{R}otation \textbf{A}daptation (\textbf{SORA}), which employs a parameter-efficient orthogonal rotation to perform a more powerful, norm-preserving transformation of the feature space. Extensive experiments on natural language understanding and generation tasks demonstrate that our proposed methods achieve superior performance and efficiency compared to both LoRA and DoRA. The code is available at \url{https://github.com/MaeChd/SORA}.
\end{abstract}

\section{Introduction}

Large-scale Pre-trained Models constitute the bedrock of modern artificial intelligence, yet their substantial fine-tuning costs drive research into Parameter-Efficient Fine-Tuning (PEFT)~\cite{ding2023parameter}. PEFT techniques adapt these models with minimal computational and storage overhead through diverse strategies, ranging from inserting small, trainable adapter modules into the model's architecture~\cite{houlsby2019parameter} to methods like prompt-tuning, which only modify the input embeddings~\cite{Lester2021ThePO,li2021prefix}. Among these, LoRA~\cite{Hu2021LoRALA} is one of the most widely adopted PEFT techniques, which operates by adding a low-rank update $\mathbf{BA}$ to the pre-trained weights $\mathbf{W_{\text{pre}}}$ (i.e., $\mathbf{W = W_{pre}} + s\cdot\mathbf{BA}$).

The simplicity and effectiveness of LoRA inspire a continuous stream of research, leading to numerous variants that improve upon the original method along several key axes. These advancements primarily focus on: \textit{(i)} dynamic rank allocation during training, as exemplified by AdaLoRA~\cite{Zhang2023AdaLoRAAB} and ReLORA~\cite{Lialin2023ReLoRAHT}; \textit{(ii)} improved initialization of low-rank matrices from pre-trained weights, proposed in methods like PiSSA~\cite{Meng2024PiSSAPS} and LoRA-GA~\cite{Wang2024LoRAGALA}; and \textit{(iii)} refined optimization strategies, such as those introduced in LoRA+~\cite{Hayou2024LoRAEL} and LoRA-RITE~\cite{yen2024lora}.

Our work connects to the thread of principled weight preparation, with a focus on DoRA (Weight-Decomposed Low-Rank Adaptation)~\cite{Liu2024DoRAWL}. As an improvement upon LoRA, DoRA pushes this philosophy further by decoupling a pre-trained weight matrix $\mathbf{W}$ into its magnitude and direction. It then applies the low-rank update exclusively to the directional component, leaving the magnitude fixed. This process is formulated as $\mathbf{W} = \mathbf{m} \frac{\mathbf{W_{\text{pre}}} + s\mathbf{BA}}{\|\mathbf{W_{\text{pre}}} + s\mathbf{BA}\|_c}$.

DoRA exhibits superior performance to LoRA on numerous tasks, demonstrating the potential of this decomposition. However, this success is accompanied by a significant drawback: the column-wise norm computation ($\|\cdot\|_c$) required during training introduces substantial computational overhead. This overhead hinders its practical application, especially for larger models and longer sequences. Concurrently, a more fundamental question remains unanswered:

\begin{tcolorbox}
\begin{center}
    \textbf{\textit{What is the true driving force behind DoRA's performance improvement?}}
\end{center}
\end{tcolorbox}

\begin{figure*}[!htbp]
    \centering
    \subfloat[Stable Rank]{\label{Fig:G1}%
    \begin{minipage}[h]{0.5\textwidth}
        \centering
        \includegraphics[width=\textwidth]{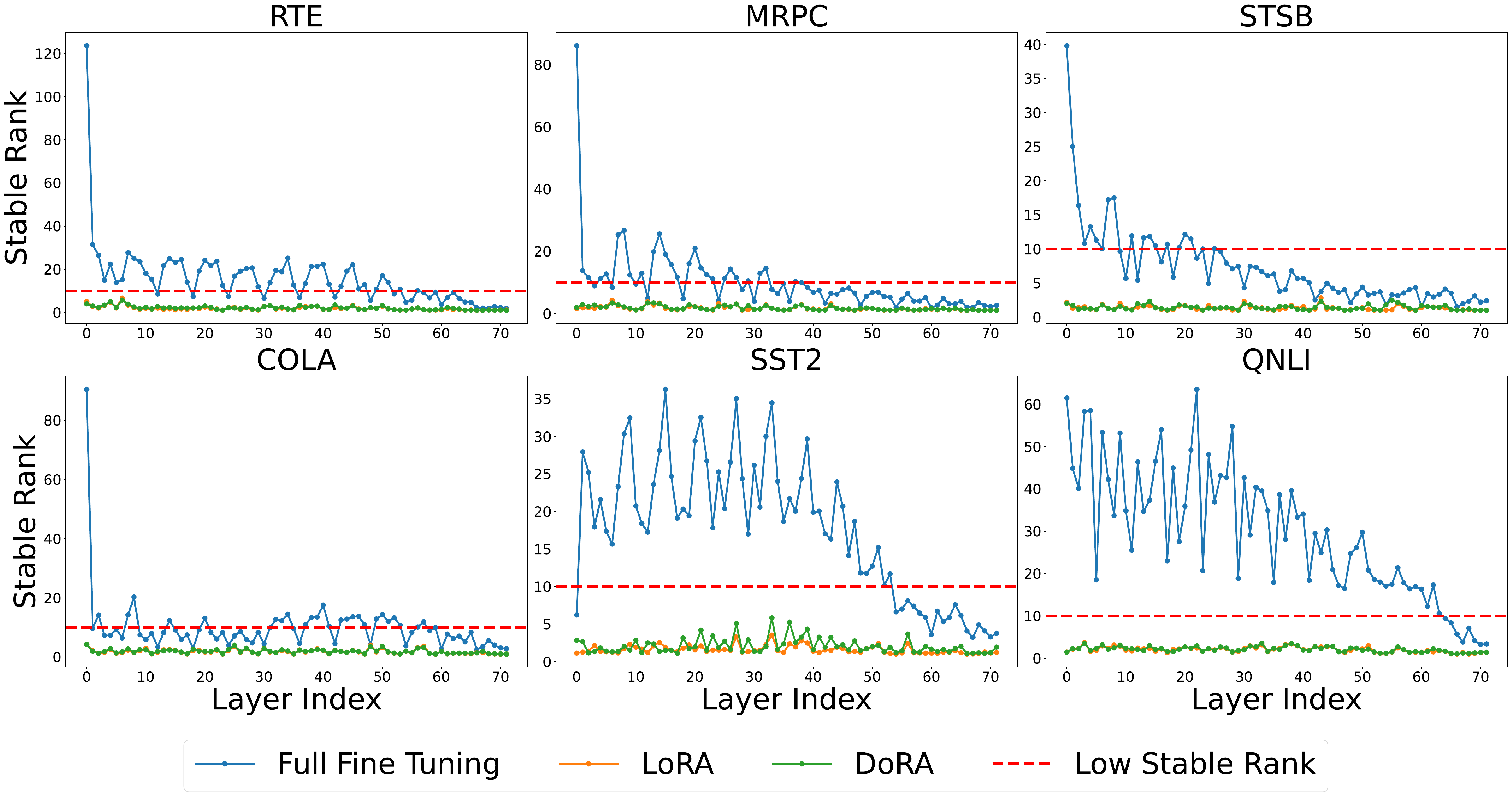}
    \end{minipage}
    }
    \subfloat[SVD Entropy]{\label{Fig:G2}%
    \begin{minipage}[h]{0.5\textwidth}
        \centering
        \includegraphics[width=\textwidth]{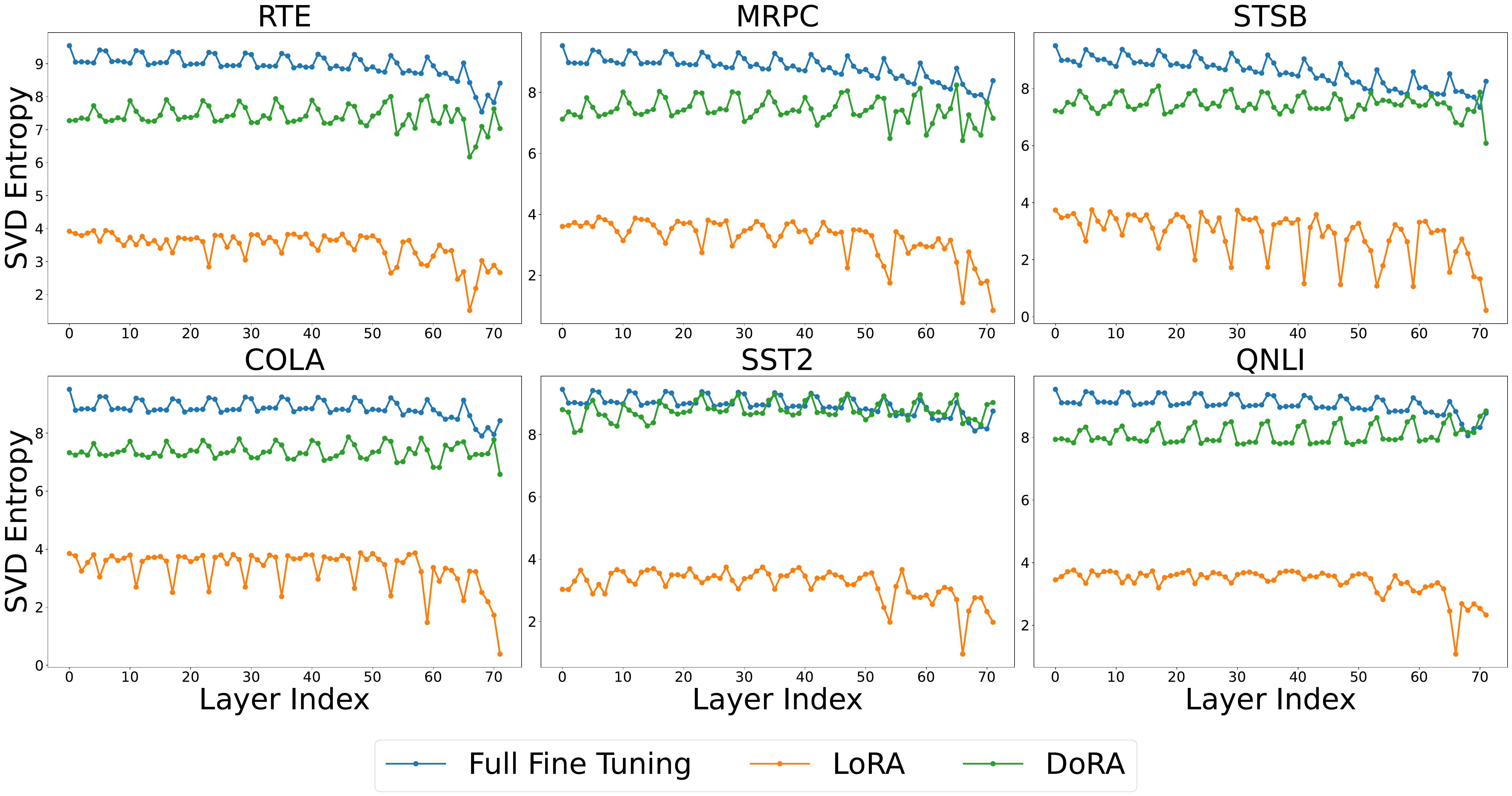}
    \end{minipage}
    }
    
    \caption{(a) DeBERTaV3-Base on GLUE Benchmark: Stable Rank $\|\Delta \mathbf{W}\|_F^2 / \|\Delta \mathbf{W}\|_2^2$ Across Layers; (b) DeBERTaV3-Base on GLUE Benchmark: SVD Entropy $H(\sigma) = -\sum_i p_i \log p_i$ Across Layers, with Comparisons of Full Fine-Tuning, LoRA, and DoRA.}
    \label{Fig:FG34}
\end{figure*}

To answer this question, we conduct an in-depth empirical analysis. We find that while prior work points to stable rank~\cite{Lion2025PoLARPL}, its explanatory power is limited in many practical tasks. 
Instead, we find that DoRA consistently increases the \textbf{Singular Value Entropy} of the weight update matrix. A higher entropy signifies a broader distribution of update energy across features, which prevents over-specialization and emulates the desirable behavior of full fine-tuning~\cite{Liu2025MuonIS}. This suggests DoRA's success stems from indirectly optimizing the weight update's singular value spectrum. To validate this and address DoRA's inefficiency, we reframe it as a more efficient \textbf{Weight Conditioning} method~\cite{Saratchandran2024WeightCF}. This perspective reveals DoRA's mechanism as modulating the main weight matrix with a lightweight, learnable matrix. This discovery opens a broader design space, for which we propose a unified weight conditioning framework to create more powerful and efficient PEFT methods.

Within this framework, we introduce two novel methods: \textit{(i)} \textbf{Pre-Diag} which efficiently calibrates pre-trained weights to improve performance while reducing training time, and \textit{(ii)} \textbf{S}kewed \textbf{O}rthogonal \textbf{R}otation \textbf{A}daptation (\textbf{SORA}), a powerful, norm-preserving transformation that uses parameter-efficient orthogonal rotations to align the feature space, building on Orthogonal Fine-Tuning (OFT)~\cite{Qiu2023ControllingTD}.

Our contributions can be summarized as follows:
\begin{enumerate}
    \item We are the first to connect the success of DoRA to an increase in singular value entropy and, based on this, reframe it as an efficient weight conditioning method.
    \item We propose a unified weight conditioning framework, from which we derive two novel methods: Pre-Diag for efficient weight calibration and SORA for powerful rotational adaptation of the feature space.
    \item We demonstrate through extensive experiments that our proposed methods outperform existing state-of-the-art PEFT methods in terms of both performance and efficiency.
\end{enumerate}

\section{Analysis and Insights}
In this section, we first demonstrates the inadequacy of stable rank for explaining performance differences among fine-tuning methods. We then introduce singular value entropy as a more discerning metric, providing both empirical and theoretical evidence that DoRA attains a higher entropy than LoRA. Ultimately, we uncover DoRA's core mechanism by reformulating it as a form of weight conditioning, which enables it to modify the full singular value spectrum and thus enhance update uniformity.
\subsection{The Inadequacy of Stable Rank}
To understand the performance discrepancies among different fine-tuning methods, we first analyze the stable rank of the parameter update matrix, $\Delta \mathbf{W}=\mathbf{W}-\mathbf{W}_{\text{pre}}$, where $\mathbf{W}$ represents the final weights and $\mathbf{W_{\text{pre}}}$ the initial pre-trained weights. The stable rank is defined as $\|\Delta \mathbf{W}\|_F^2 / \|\Delta \mathbf{W}\|_2^2$~\cite{rudelson2007sampling}. Our choice of this metric is motivated by prior work that has linked it to the diversity of update directions~\cite{Lion2025PoLARPL} and the expressive capacity of low-rank models~\cite{Zeng2023TheEP,Kalajdzievski2023ARS}. Our initial hypothesis was that the stable rank could quantitatively explain why methods like Full Fine-Tuning (FFT) and DoRA outperform LoRA.

However, our layer-wise analysis reveals that the aggregated, model-wide stable rank is an insufficient and potentially misleading metric. 
As shown in Figure \ref{Fig:G1}, the stable rank of FFT is highly volatile across layers, soaring in some while dropping to levels comparable to the consistently low ranks of LoRA and DoRA in others. This volatility indicates that FFT's high global rank is driven by only a subset of layers. Consequently, stable rank fails to robustly distinguish the learning dynamics of these methods or explain their performance differences.

\begin{figure}[!htbp]
	\centering
    \includegraphics[width=0.45\textwidth]{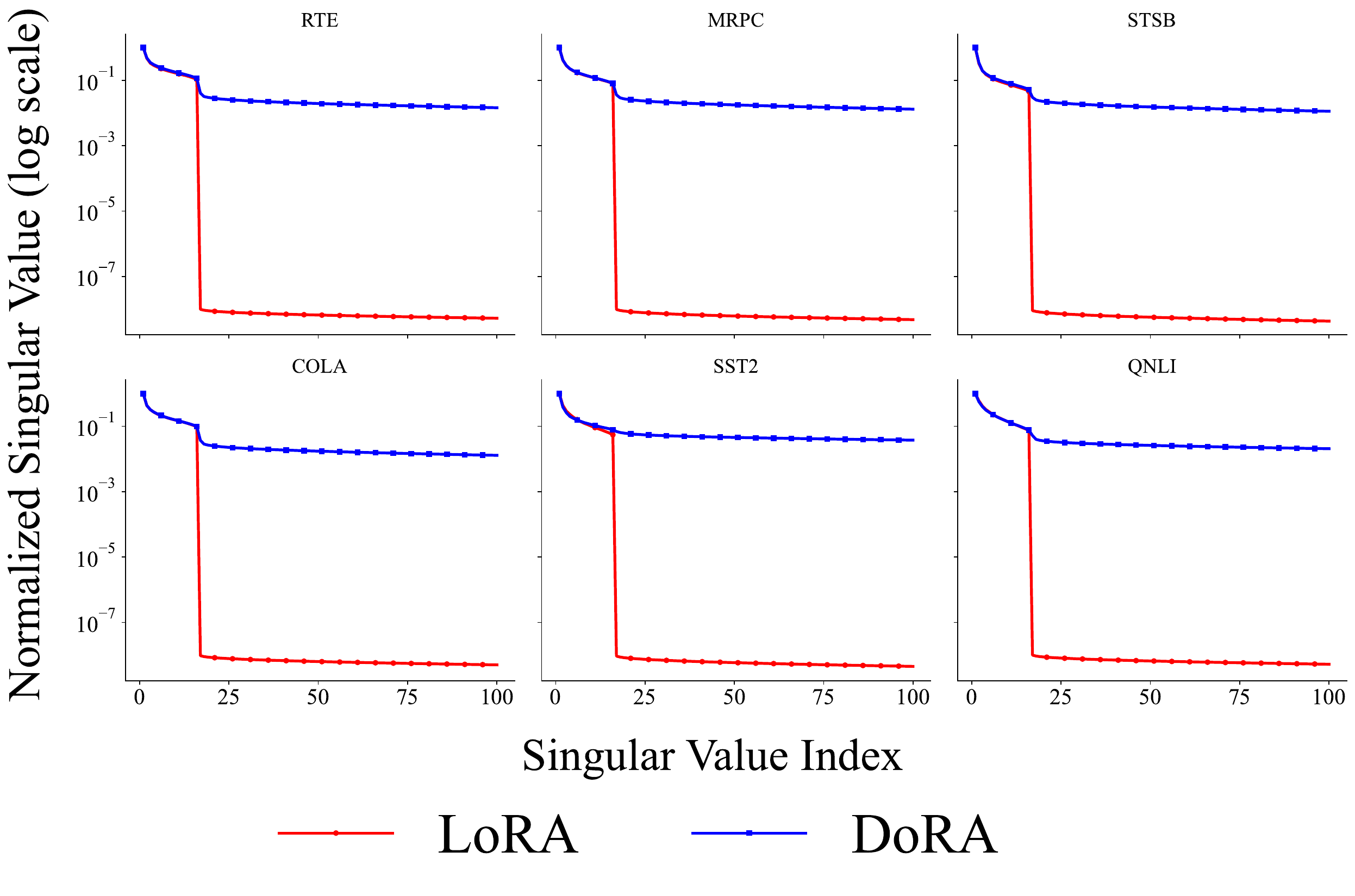}
    \caption{Distribution of average singular values after layer-wise normalization for LoRA and DoRA on GLUE tasks with DeBERTaV3-Base.}
    \label{fig:normsvd}
\end{figure}

\subsection{SVD Entropy as a More Faithful Explanator}

The limitations of stable rank motivate our search for a more discerning metric. We therefore employ singular value entropy, defined as the Shannon entropy of the normalized singular value spectrum: $H(\sigma) = -\sum_i p_i \log p_i$, where $p_i = \sigma_i^2 / \sum_j \sigma_j^2$ ~\cite{7098875,doi:10.1073/pnas.97.18.10101}. Intuitively, this metric measures how uniformly the update's energy is distributed across its singular directions. A higher entropy indicates that the update is more uniformly distributed, rather than being dominated by a few high-magnitude singular values. This uniformity prevents the update from concentrating on only a few dominant features.

In stark contrast to the ambiguous results from stable rank, Figure \ref{Fig:G2} shows that layer-wise singular value entropy reveals a clear and consistent hierarchy: FFT consistently maintains the highest entropy, followed closely by DoRA, with LoRA lagging significantly behind.

To provide theoretical support for this empirical finding, we analyze a simplified model inspired by the step-like structure of the singular value distributions observed in Figure \ref{fig:normsvd}. We model the normalized singular values of LoRA and DoRA as two-step and three-step distributions, respectively. This leads to the following proposition:

\begin{theorem}
\label{th:1}
Let $\Delta \mathbf{W}_{\text{LoRA}}$ be a LoRA update matrix of rank $r$. Its set of normalized singular values is $\Sigma_{\text{LoRA}} = \{1, \underbrace{\alpha, \dots, \alpha}_{r-1}\}$. Let $\Delta \mathbf{W}_{\text{DoRA}}$ be a DoRA update matrix of rank $s > r$. Its set of normalized singular values is $\Sigma_{\text{DoRA}} = \{1, \underbrace{\beta, \dots, \beta}_{r-1}, \underbrace{\gamma, \dots, \gamma}_{s-r}\}$. The singular value entropy $H(\sigma(\Delta \mathbf{W}))$ is the Shannon entropy of the probability distribution $\{p_i\}$ formed by the squared normalized singular values, where $p_i = \sigma_i^2 / \sum_j \sigma_j^2$. Given the constraint that the sum of squares of minor singular values is preserved, $(r-1)\alpha^2 = (r-1)\beta^2 + (s-r)\gamma^2$, and assuming $\alpha > \gamma > 0$ and $\beta > 0$. Then, the singular value entropy of the DoRA matrix is strictly greater than that of the LoRA matrix:
$$H(\sigma(\Delta \mathbf{W}_{\text{DoRA}})) > H(\sigma(\Delta \mathbf{W}_{\text{LoRA}}))$$
\end{theorem}
A detailed proof is provided in the Appendix 1. 

\begin{remark}
    This proposition formally demonstrates that by activating more singular directions and distributing the update energy across them (even if the newly activated values $\gamma$ are small), the overall entropy of the update matrix increases. This reinforces our central finding: the ability to increase singular value entropy appears to be the primary driving force behind DoRA's performance gains over LoRA.
\end{remark}

\subsection{Unveiling DoRA: Weight Conditioning}
This finding raises a crucial question: \textbf{how does the DoRA formulation mechanically increase singular value entropy?} The original DoRA update rule~\cite{Liu2024DoRAWL} decomposes the weight matrix $\mathbf{W}$ into a learnable magnitude $\mathbf{m}$ and a direction component:
$$
\mathbf{W} = \mathbf{m} \frac{\mathbf{W_{\text{pre}}}+s\cdot\mathbf{BA}}{\|\mathbf{W_{\text{pre}}}+s\cdot\mathbf{BA}\|_c},
$$
where $\|\cdot\|_c$ denotes the column norm.

We can algebraically reformulate this operation as a form of weight conditioning~\cite{Saratchandran2024WeightCF}. This reformulation reveals the effective weight change $\Delta \mathbf{W}_{\text{DoRA}} = \mathbf{W} - \mathbf{W_{\text{pre}}}$ to be:
$$
\Delta \mathbf{W}_{\text{DoRA}} = \mathbf{W_{\text{pre}}}(\mathbf{D}-\mathbf{I}) + s\cdot\mathbf{BA}\mathbf{D},
$$
where $\mathbf{D} = \text{Diag}\left(\frac{\mathbf{m}}{\|\mathbf{W_{\text{pre}}}+s\cdot\mathbf{BA}\|_c}\right)$.

This reformulation offers two critical insights. Computationally, it replaces the costly column-wise norm with an efficient matrix multiplication. More importantly, it reveals DoRA's core mechanism: unlike the strictly low-rank LoRA update, the DoRA update $\Delta \mathbf{W}_{\text{DoRA}}$ incorporates a higher-rank term, $\mathbf{W_{\text{pre}}}(\mathbf{D}-\mathbf{I})$.

This higher-rank component is key. For singular value indices $i > r$, the LoRA update is zero (i.e., $\sigma_i(\Delta \mathbf{W}_{\text{LoRA}}) = 0$). In contrast, Weyl's theorem~\cite{horn2012matrix} shows that DoRA maintains non-zero singular values in these dimensions:
$$
\sigma_i(\Delta \mathbf{W}_{\text{DoRA}}) \leq \sigma_1(\mathbf{W_{\text{pre}}}(\mathbf{D}-\mathbf{I})) + \sigma_i(\mathbf{BAD}) \quad \text{for } i > r.
$$
This ability to modify the full singular value spectrum allows DoRA to create a richer, more uniform update distribution, thereby increasing its entropy, as illustrated in Figure \ref{fig:normsvd}.

\begin{table*}[!h]
\caption{Comparison of properties between Fully Fine-Tuning, LoRA, DoRA and our's Pre-Diag, SORA.}
\label{tb:compare}
\begin{center}
\begin{small}
\begin{tabular}{lcccc}
\toprule
\textbf{Method} &\makecell{\textbf{Formulation}} &\makecell{\textbf{Condition}}  & \makecell{\textbf{Placement}} &
\makecell{\textbf{Key idea}} \\
\midrule
LoRA~\cite{Hu2021LoRALA}  & $\mathbf{W_{\text{pre}}+BA}$  & Identity Matrix & N/A & Low-Rank Adaption\\

DoRA~\cite{Liu2024DoRAWL} & $\mathbf{(W_{\text{pre}}+BA)D}$ & Diagonal Matrix & Post & Weight-Decomposed\\

\textbf{Pre-Diag} (ours) & $\mathbf{W_{\text{pre}}D+BA}$ & Diagonal Matrix  & Pre & Calibration\\

\textbf{SORA} (ours) &$\mathbf{(W_{\text{pre}}D+BA)P}$ & Orthogonal Matrix & Pre \& Post &  Calibration \&  Rotation\\
\bottomrule
\end{tabular}
\end{small}
\end{center}
\end{table*}

\begin{figure*}[!htbp]
	\centering
    \includegraphics[width=0.8\textwidth]{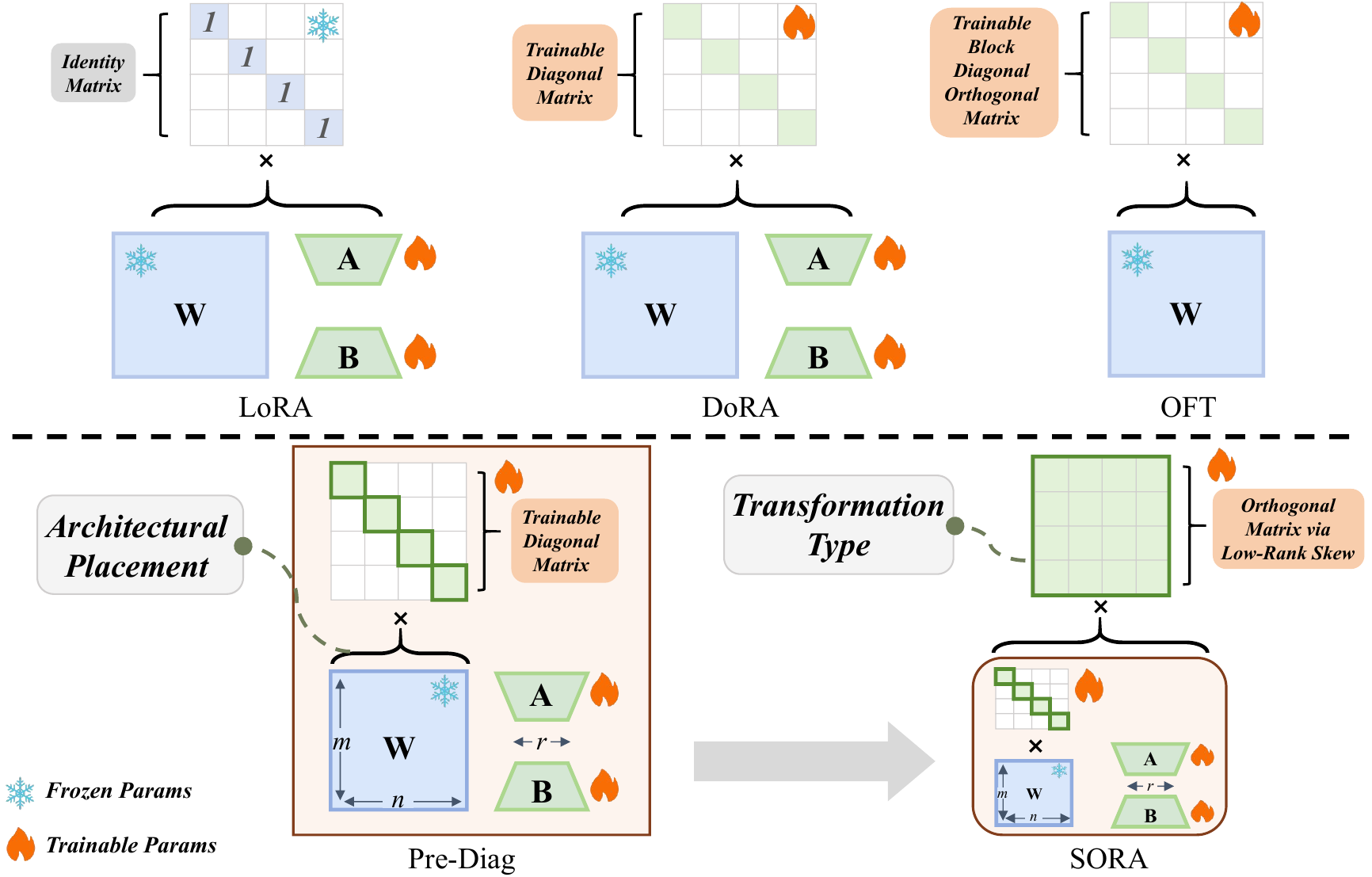}  
    \caption{Taking LoRA as the baseline, we can regard DoRA and OFT as weight-conditioned methods built upon it. Our method Pre-Diag, in turn, adjusts pre-trained weights by using a trainable diagonal matrix based on DoRA; SORA, on the other hand, replaces the conditional matrix in DoRA with a trainable low-parameter orthogonal matrix to automatically rotate input features.}
    \label{fig:method}
\end{figure*}

\section{Framework and Methods}
Our analysis reveals that DoRA's success stems from its capacity to increase the singular value entropy of the update matrix. This effect is achieved through an implicit form of weight conditioning, where a diagonal matrix $\mathbf{D}$ modulates the weights. Based on this insight, we propose a Unified Framework for Weight Conditioning to move beyond ad-hoc improvements and systematically design novel PEFT methods.

The framework's core idea is to enhance the standard LoRA update, $\Delta \mathbf{W} = s\mathbf{BA}$, with a learnable conditioning matrix. We identify two fundamental and orthogonal design axes:
\begin{enumerate}
    \item  \textbf{Architectural Placement}: Where should the conditioning matrix be applied? Should it act on the entire weight after the LoRA update as in DoRA, or only on the pre-trained weights?
    \item  \textbf{Transformation Type} : What transformation should the conditioning matrix perform? Should it be a diagonal matrix $\mathbf{D}$ that only performs axis-aligned scaling as implicitly done by DoRA, or a more powerful orthogonal matrix $\mathbf{P}$ capable of performing rotation?
\end{enumerate}
By systematically exploring these two axes, we develop principled and efficient methods that explicitly optimize the update spectrum.

\subsection{Placement: Pre-Diag for Efficient Calibration}

We first explore the architectural placement of the conditioning matrix. Recall from previous section that DoRA's update can be expressed as $\Delta \mathbf{W}_{\text{DoRA}} = \mathbf{W_{\text{pre}}}(\mathbf{D}-\mathbf{I}) + s\cdot\mathbf{BA}\mathbf{D}$. In this formulation, the conditioning matrix $\mathbf{D}$ is coupled with both the pre-trained weights $\mathbf{W_{\text{pre}}}$ and the LoRA update $\mathbf{BA}$, creating a complex gradient path.

To simplify this, we propose Pre-Diag, which applies the conditioning directly and solely to the pre-trained weights. Its formulation is:
$$\mathbf{W} = \mathbf{W_{\text{pre}}}\mathbf{D} + s\cdot\mathbf{BA}.$$
The resulting update matrix is $\Delta \mathbf{W}_{\text{Pre-Diag}} = \mathbf{W_{\text{pre}}}(\mathbf{D}-\mathbf{I}) + s\cdot\mathbf{BA}$.

The motivation for this design is to decouple two distinct goals: (1) using the diagonal matrix $\mathbf{D}$ to efficiently calibrate the scale of the pre-trained weights $\mathbf{W_{\text{pre}}}$, and (2) using the LoRA update $\mathbf{BA}$ to learn new, task-specific features. This cleaner, more direct structure simplifies the gradient computation and, as we show in experiments, leads to faster training. Pre-Diag thus serves as a more direct and efficient method for spectrum optimization, achieving high singular value entropy with reduced computational overhead.

\subsection{Transformation: SORA for Rotational Adaptation}
We further enhance the Pre-Diag, $\mathbf{W} = \mathbf{W_{\text{pre}}}\mathbf{D} + s\cdot\mathbf{BA}$, by introducing a learnable transformation. This model implicitly ends with an identity matrix: $(\mathbf{W_{\text{pre}}}\mathbf{D} + s\cdot\mathbf{BA})\mathbf{I}$. We propose replacing this static identity with a dynamic, parameter-efficient rotation matrix $\mathbf{P}$.

This leads to our method, \textbf{SORA} (\textbf{S}kewed \textbf{O}rthogonal \textbf{R}otation \textbf{A}daptation), which applies this rotation to refine feature interactions:
$$\mathbf{W}' = (\mathbf{W_{\text{pre}}}\mathbf{D} + s\cdot\mathbf{BA})\mathbf{P}.$$
To maintain efficiency, the orthogonal matrix $\mathbf{P}$ is constructed as the exponential of a low-rank skew-symmetric matrix $\mathbf{S_P}$, where $\mathbf{S_P}$ is parameterized by two small matrices $\mathbf{D_P}, \mathbf{C_P} \in \mathbb{R}^{n \times r_P}$ ($r_P \ll n$).

As the direct computation of the matrix exponential, $\mathbf{P} = \exp(s_P \cdot\mathbf{S_P})$, is prohibitively expensive, we employ a highly accurate first-order Taylor approximation for small rotations:
$$\mathbf{P} \approx \mathbf{I} + s_P\cdot \mathbf{S_P} = \mathbf{I} + s_P\cdot (\mathbf{D_P C_P}^\top - \mathbf{C_P D_P}^\top).$$
This approximation makes $\mathbf{P}$ near-orthogonal while drastically reducing computational demands. \cite{Singla2021SkewOC} first introduced the skew-orthogonal strategy into CNNs and provided a bounding for the approximation of matrix exponentials. Similarly, in SORA, for the orthogonal matrix $\mathbf{P}$, we have the following theorem:

\begin{theorem}
\label{th:2}
For Skew-Symmetric $\mathbf{J}$, we have the inequality:
$$
\|\exp(\mathbf{J})-T_k(\mathbf{J})\|_2\leq\frac{\|\mathbf{J}\|_2^k}{k!}\quad\text{where}\quad T_k(\mathbf{J})=\sum_{i=0}^{k-1}\frac{\mathbf{J}^i}{i!}.
$$
Then, Let $s_P = \frac{\epsilon}{2\|\mathbf{D_P}\|_F\|\mathbf{C_P}\|_F + \epsilon}$, we have the inequality:
$$
\|\exp(\mathbf{P}) - T_1(\mathbf{P})\|_2\le \epsilon.
$$
\end{theorem}
A detailed proof is provided in the Appendix 2.
\begin{remark}
This implies that when $\|\mathbf{P}\|_F$ is extremely small, $\exp(\mathbf{P})$ can be approximated with high precision, and this can be achieved by controlling $s_P$, thereby significantly reducing the time complexity involved in solving $\exp(\mathbf{P})$.
\end{remark}
To efficiently compute $\mathbf{W_{PD}} \mathbf{P}$ where $\mathbf{W_{PD}} = \mathbf{W_{\text{pre}}D} + s\cdot\mathbf{BA}$, we optimize the matrix multiplication. A naive calculation of $\mathbf{W_{PD}}(\mathbf{D_P C_P}^\top - \mathbf{C_P D_P}^\top)$ has $\mathcal{O}(mn^2)$ complexity. By reordering the operations as $(\mathbf{W_{PD}D_P})\mathbf{C_P}^\top - (\mathbf{W_{PD}C_P})\mathbf{D_P}^\top$, we reduce the complexity to $\mathcal{O}(mnr_P)$, ensuring SORA is both powerful and computationally efficient.

Figure \ref{fig:method} and Table \ref{tb:compare} detail the differences between our proposed \textbf{Pre-Diag} and \textbf{SORA} and existing methods.

\section{Experiment}
In this section, we evaluate the performance of our proposed methods, Pre-Diag and SORA.
\paragraph{Setup} All experiments were conducted on a setup consisting of four NVIDIA RTX 4090 (24GB) GPUs and eight Ascend 910C (64GB) NPUs. Without additional specifications, we use $r_P = 1$ by default in our experiments. Detailed experimental settings are provided in Appendix 3.

\paragraph{Natural language understanding}
To test general language understanding, we fine-tuned DeBERTaV3-Base \cite{He2020DeBERTaDB} on the GLUE benchmark\cite{Wang2018GLUEAM}. 
We test three learning rates ($\{1\text{e-}3,8\text{e-}4,4\text{e-}4\}$), each with three random seeds for robustness. Table \ref{table-glue} shows aggregated results comparing our Pre-Diag and SORA with LoRA and DoRA, reporting the best average performance. The findings indicate that our methods achieve superior average performance on the benchmark. 

\begin{table*}[!h]
\caption{Comparison of best average results of fine-tuning DeBERTaV3-base model on GLUE benchmark. We report the Matthew’s (Pearson) correlation for CoLA (STS-B), and accuracy for other tasks}
\label{table-glue}
\begin{center}
\begin{small}
\begin{tabular}{cccccccccc}
\toprule
\textbf{Method} &\makecell{\textbf{Trainable} \\ \textbf{Params Rate}} & \makecell{\textbf{RTE}\\2.5k} & \makecell{\textbf{MRPC}\\3.7k} & \makecell{\textbf{STS-B}\\7k} & 
\makecell{\textbf{CoLA}\\8.5k}  & 
\makecell{\textbf{SST-2}\\67k} & \makecell{\textbf{QNLI}\\105k} &\makecell{\textbf{QQP}\\364k}  & \textbf{Avg.} \\
\midrule

LoRA & 1.42\% & 86.63$_{\pm0.63}$ & 91.25$_{\pm0.79}$ & 90.44$_{\pm0.36}$ &  69.71$_{\pm0.51}$ & 96.21$_{\pm0.00}$ &  93.94$_{\pm0.10}$ & 91.80$_{\pm0.60}$ & 88.57 \\
DoRA & 1.46\% & 86.83$_{\pm1.02}$ & 92.15$_{\pm0.73}$ & 90.34$_{\pm0.41}$ & 69.90$_{\pm0.45}$  & 96.25$_{\pm0.07}$ & 94.06$_{\pm0.27}$ & 92.38$_{\pm0.02}$ & 88.84 \\

\rowcolor{customgray}\textbf{Pre-Diag} & \textbf{1.46\%} & 87.24$_{\pm1.67}$  & 91.17$_{\pm0.25}$ & 90.50$_{\pm0.46}$ & 71.19$_{\pm0.49}$ & 96.10$_{\pm0.20}$ & 94.04$_{\pm0.14}$ & 92.21$_{\pm0.03}$ &  \textbf{88.98}  \\

\rowcolor{customgray}\textbf{SORA} & \textbf{1.55\%}& 87.90$_{\pm0.42}$  & 91.82$_{\pm1.02}$ & 90.79$_{\pm0.57}$ & 70.80$_{\pm1.18}$ & 96.36$_{\pm0.12}$ & 94.30$_{\pm0.23}$ & 92.38$_{\pm0.06}$ &  \textbf{89.19}  \\

\bottomrule
\end{tabular}
\end{small}
\end{center}
\end{table*}

\begin{figure}[!htbp]
	\centering
    \includegraphics[width=0.48\textwidth]{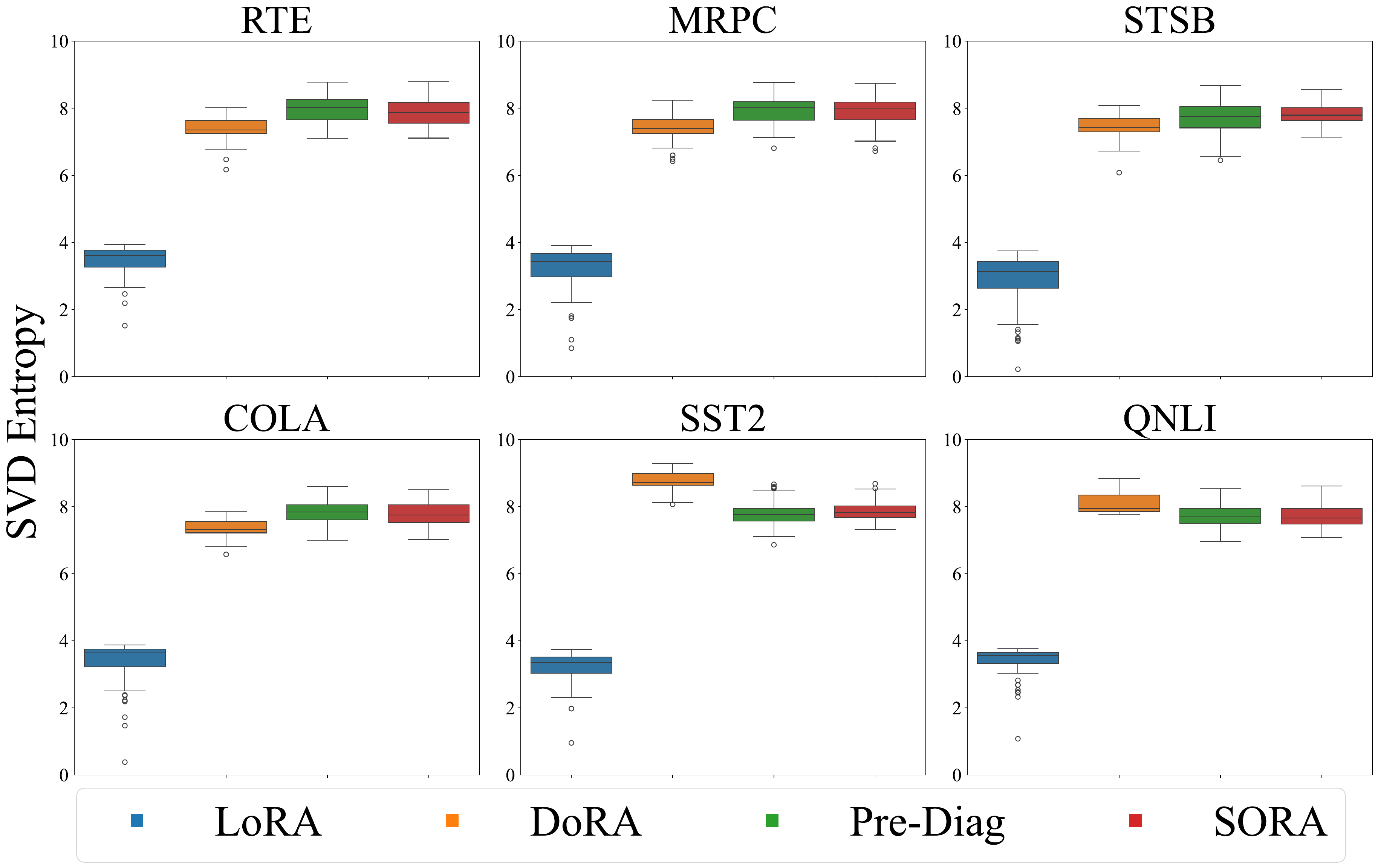}   
    \caption{SVD Entropy of DeBERTaV3-Base Fine-Tuned on GLUE tasks: Comparing LoRA, DoRA, Our Pre-Diag, and SORA.}
    \label{fig:boxplot}
\end{figure}

\paragraph{Commonsense reasoning}
For commonsense reasoning, we conduct several experiments. First, we train LLaMA3-8B \cite{Dubey2024TheL3} on the Commonsense15k dataset and test its performance on a commonsense benchmark.  The experimental results are presented in Table \ref{table-commonsense}. SORA achieves the optimal average score across all tasks with different ranks, which effectively confirms the effectiveness of SORA.

\begin{table*}[!h]
\caption{Single-task commonsense reasoning performance on LLaMA3-8B. Models are trained on commonsense15k, comparing LoRA, DoRA, and our method across a range of ranks.}
\label{table-commonsense}
\begin{center}
\begin{small}
\begin{tabular}{l c c c c c c c c c c}
\toprule
\makecell{\textbf{Rank}} & \textbf{Method} &  \makecell{\textbf{BoolQ}} & \makecell{\textbf{PIQA}} & \makecell{\textbf{SIQA}} & \makecell{\textbf{HeSw}} & \makecell{\textbf{WiGr}} & \makecell{\textbf{ARC-e}} & \makecell{\textbf{ARC-c}} & \makecell{\textbf{OBQA}} & \textbf{Avg.} \\
\midrule
\multirow{4}{*}{4} & LoRA &  68.96 & 75.46 & 73.18 &  79.40 & 71.19 & 84.09 & 69.54 & 78.20 & 75.00 \\
 & DoRA & 69.72 & 75.35 & 72.31 & 78.36 & 73.88  & 83.92 & 69.45 & 78.20 & 75.15 \\

\rowcolor{customgray}
\cellcolor{white}\multirow{-2}{*}{4}& \textbf{Pre-Diag} &  69.02 & 77.09 & 73.08 & 79.27 & 71.59 & 80.98 & 68.94 & 74.20 & 74.27 \\

\rowcolor{customgray}
\cellcolor{white}& \textbf{SORA} & 69.63 & 76.28 & 72.62 & 78.92 & 71.59 & 84.43 & 70.90 & 77.80 & \textbf{75.27} \\

\midrule
\multirow{4}{*}{8} & LoRA &  71.13 & 74.76 & 72.52 & 81.49 & 74.19 & 84.22 & 70.31 & 77.60 & 75.78\\
& DoRA &  70.86 & 75.90 & 72.72 & 80.85 & 74.90 & 84.89 & 70.73 & 78.80 & 76.08 \\

\rowcolor{customgray}
\cellcolor{white}\multirow{-2}{*}{8}& \textbf{Pre-Diag} &  70.06 & 77.09 & 73.08 & 81.68 & 72.61 & 84.68 & 70.56 & 79.40 & \textbf{76.15}\\

\rowcolor{customgray}
\cellcolor{white}& \textbf{SORA} & 70.76 & 76.66 & 73.03 & 81.49 & 73.40 & 84.89 & 70.73 & 78.80 & \textbf{76.22}\\
\midrule

\multirow{4}{*}{16}  & LoRA & 70.09 & 76.01 & 73.59 & 81.08 & 73.88 & 84.60 & 70.90 & 78.80 & 76.12 \\
& DoRA &  71.77 & 76.33 & 72.76 & 82.27 & 74.19 & 83.84 & 68.77 & 79.60 & 76.19\\

\rowcolor{customgray}
\cellcolor{white}\multirow{-2}{*}{16}& \textbf{Pre-Diag} &  70.24 & 76.71 & 73.44 & 81.26 & 74.82 & 84.64& 71.16 & 79.40 & \textbf{76.46}\\

\rowcolor{customgray}
\cellcolor{white}&\textbf{SORA} &  71.56 & 76.71 & 73.69 & 82.82 & 76.40 & 85.43 & 70.14 & 79.00 & \textbf{76.97}\\
\midrule
\multirow{4}{*}{32} & LoRA &  71.87 & 77.97 & 73.85 & 83.46 & 76.40 & 85.98 & 71.33 & 80.20 & 77.63 \\
& DoRA &  71.35 & 78.56 & 73.29 & 83.70 & 75.92 & 86.03 & 71.67 & 80.40 & 77.62\\

\rowcolor{customgray}
\cellcolor{white}\multirow{-2}{*}{32}& \textbf{Pre-Diag} &  71.83 & 78.07 & 73.03 & 83.33 & 75.37 & 85.48 & 72.44 & 81.20 & \textbf{77.59}\\

\rowcolor{customgray}
\cellcolor{white}& \textbf{SORA} & 72.32 & 78.67 & 73.95 & 83.77 & 76.87 & 86.53 & 71.93 & 80.40 & \textbf{78.06}  \\
 
\bottomrule
\end{tabular}

\end{small}
\end{center}
\end{table*}

\begin{figure}[!h]
\centering
    \includegraphics[width=0.48\textwidth]{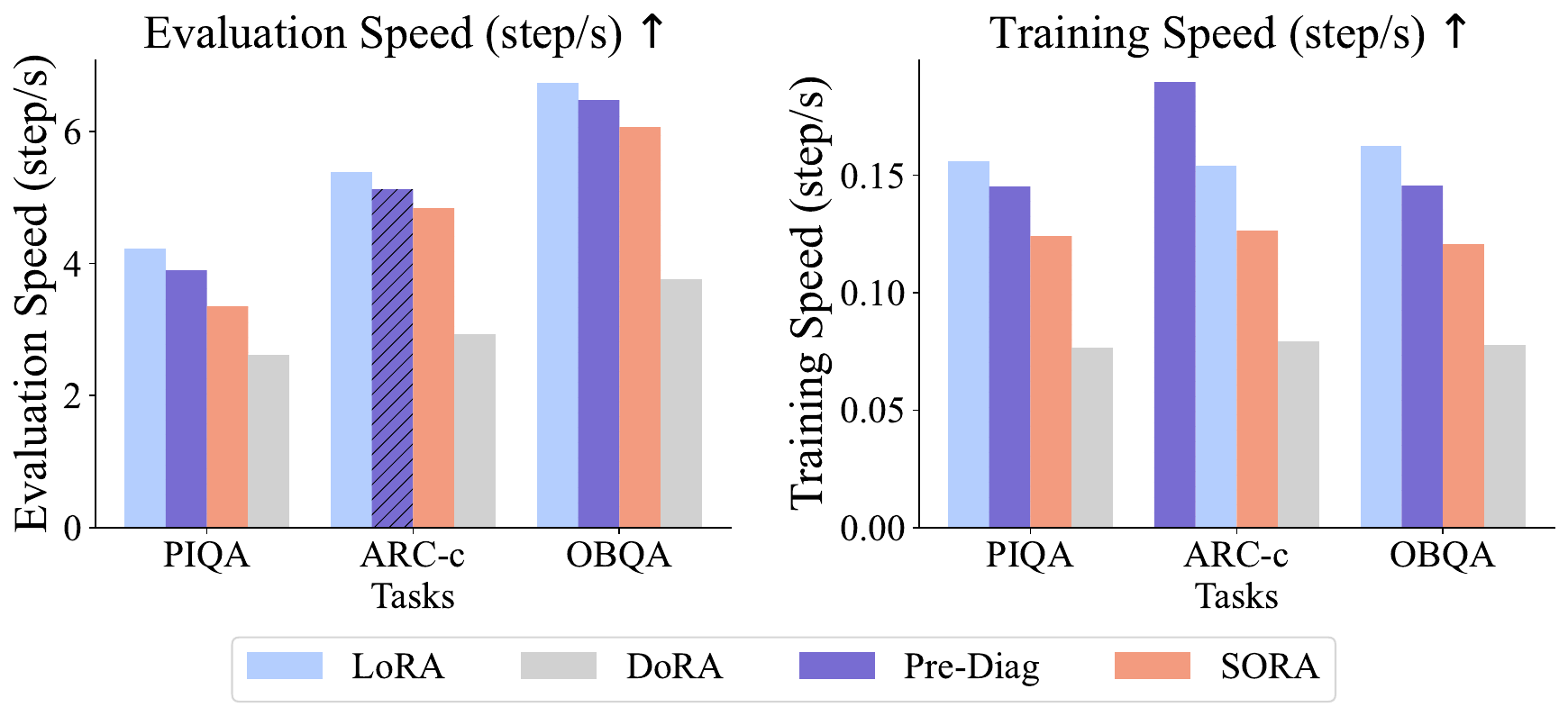}   
    \caption{ Comparison of inference speed (steps per second) and training speed (steps per second) among LoRA, DoRA, and our proposed method on the PIQA, ARC-c, and OBQA tasks, using the LLaMA3-8B model. 
    }
    \label{fig:speed}
\end{figure}


\paragraph{Mathematical reasoning} For mathematical reasoning, we conduct several experiments. First, we train Gemma-7B \cite{gemma_2024} on the MetaMathQA14k dataset \cite{yu2023metamath} and test its performance on a mathematical benchmark.  The experimental results are presented in Table \ref{table-math}. SORA achieves the optimal average score across all tasks, which effectively confirms the effectiveness of SORA.

\begin{table*}[!h]
\caption{Single-task mathematical reasoning performance on Gemma-7B. Models are trained on MetaMathQA14k and evaluated on various benchmarks. All methods use a rank of 16 ($r_p=2$ for SORA).}
\label{table-math}
\begin{center}
\begin{small}
\begin{tabular}{ccccccccc}
\toprule
\textbf{Method} &\makecell{\textbf{Trainable} \\ \textbf{Params Rate}} & \makecell{\textbf{GSM8K}} & \makecell{\textbf{MultiArith}} & \makecell{\textbf{AQuA}} & 
\makecell{\textbf{SVAMP}}  & 
\makecell{\textbf{AddSub}} & \makecell{\textbf{SingleEq}}  & \textbf{Avg.} \\
\midrule

OFT & 0.58\% & 73.09 & \textbf{99.00} & 38.19 & 75.00 & 86.10 & 94.69 & 77.68 \\

LoRA+ & 0.40\% & 74.83 & \underline{98.67} & 36.58 & 75.10 & 85.32 & 93.11 & 77.27   \\
LoRA & 0.40\% & 74.07 & 97.50 & 37.80 & 75.90 & 85.08 & \underline{95.85} & 77.70  \\
DoRA & 0.41\% & 73.77 & 97.17 & \underline{38.34} & \textbf{77.8} & \textbf{88.40} & 93.92 & 78.23 \\

\rowcolor{customgray}\textbf{SORA} & 0.47\% & \textbf{74.98} & 98.50 & \textbf{40.49} & \underline{76.30} & \underline{87.63} & \textbf{95.87} & \textbf{78.96} \\

\bottomrule
\end{tabular}
\end{small}
\end{center}
\end{table*}

\subsection{Update Diversity and Efficiency}

We evaluate our methods on two fronts: update diversity and computational efficiency. \textit{(i)} Figure \ref{fig:boxplot} shows the SVD entropy of weight updates on GLUE tasks. Both Pre-Diag and SORA consistently achieve higher entropy than LoRA, and surpass DoRA on most tasks, indicating a more uniform update distribution. \textit{(ii)}We benchmark training and inference speeds on LLaMA3-8B in Figure \ref{fig:speed}. While LoRA is the fastest, DoRA introduces substantial overhead, our methods offer a compelling trade-off. Compared to DoRA,Pre-Diag and SORA significantly accelerate both training (by 50.60\% and 36.96\%) and inference (by 65.09\% and 51.27\%). These results demonstrate that our proposed methods successfully balance enhanced update diversity with high computational efficiency.

\subsection{Ablation Study}

\paragraph{Component Analysis} To validate our architectural choices, we ablate SORA along two orthogonal axes: whether the weight conditioning is placed before or after the LoRA update, and whether its transformation is diagonal or orthogonal. We benchmark four models: the LoRA baseline, the post-diagonal model DoRA, our pre-diagonal model Pre-Diag, and our final model SORA, which combines a pre-diagonal with a post-orthogonal transform. This comparison isolates the performance contribution of each component, empirically justifying the benefits of pre-conditioning and introducing an orthogonal transformation. From Table \ref{tab:ablation}, we find that Pre-Ortho leads to worse performance with adjusted LoRA weights BA, indicating rotation transformation is more suitable for global use.

\begin{table}[h!]
\centering
\caption{Ablation study of the weight conditioning framework. Pre/Post refers to the placement of the conditioning matrix. \textbf{D} and \textbf{P} denote diagonal and orthogonal transformations, respectively. SORA (\textit{ours}) applies Pre-Diag calibration followed by Post-Ortho rotation.}
\label{tab:ablation}
\resizebox{0.48\textwidth}{!}{%
\begin{tabular}{lccccc}
\toprule
\multirow{2.5}{*}{Model} & \multicolumn{2}{c}{Placement} & \multicolumn{2}{c}{Type} & \multirow{2.5}{*}{\makecell{Avg. GLUE \\Score}}
\\
\cmidrule(lr){2-3} \cmidrule(lr){4-5} 
& Pre & Post & Diag ($\mathbf{D}$) & Ortho ($\mathbf{P}$) &  \\
\midrule
\multicolumn{6}{l}{\textit{Baseline Experiments}} \\
LoRA  & & & & & 88.57 \\

DoRA & & \checkmark & \checkmark & & 88.84  \\

Pre-Diag (\textit{ours}) & \checkmark & & \checkmark & &  88.98\\
\midrule

\multicolumn{6}{l}{\textit{New Ablation Experiments}} \\

Pre-Ortho & \checkmark & & & \checkmark & 87.76 \\

Post-Ortho & & \checkmark & & \checkmark & 88.93 \\
\midrule

SORA (\textit{ours}) & \checkmark & \checkmark & \checkmark & \checkmark &  89.23\\
\bottomrule
\end{tabular}
}
\end{table}
\paragraph{Hyperparameter Sensitivity} We analyze the model's sensitivity to the rotation rank $r_P$, which controls the trade-off between expressive power and computational cost. Evaluating on the RTE task with $r_P \in \{1, 2, 4, 8, 16\}$, we find that $r_P=1$ provides an optimal balance of performance against efficiency. The results in Table \ref{tab:rp_sensitivity} confirm that this minimal rank is sufficient.

\begin{table}[h!]
\centering
\caption{Sensitivity analysis of rotational rank $r_p$ in SORA.}
\label{tab:rp_sensitivity}
\resizebox{0.45\textwidth}{!}{
\begin{tabular}{cccc}
\toprule
\makecell{Rotational \\Rank ($\boldsymbol{r_P}$)} & \makecell{Trainable \\Params} & \makecell{Avg. RTE\\ Score} & \makecell{Training Speed\\(step/s)}  \\
\midrule

1 & 1.55\% & 87.90 &  2.17\\
2 & 1.64\% & 87.72 &  1.92\\
4 & 1.81\% & 88.44 &  1.82\\
8 & 2.16\% & 88.44 &  1.75\\
16 & 2.85\% & 89.53 &  1.72\\
\bottomrule
\end{tabular}
}
\end{table}

\section{Related Work}

Parameter-Efficient Fine-Tuning (PEFT) methods reduce the significant computational and storage costs associated with adapting large pre-trained models. Among various approaches, such as adapter modules~\cite{houlsby2019parameter} and prompt-tuning~\cite{Lester2021ThePO}, methods based on low-rank adaptation are particularly prominent for their effectiveness and efficiency.

LoRA~\cite{Hu2021LoRALA} is a cornerstone of this approach. It approximates weight updates as the product of two low-rank matrices, based on the assumption that these updates occupy a low-dimensional manifold. This method is also provably capable of approximating any target transformer model under mild assumptions~\cite{Zeng2023TheEP}. This elegant mechanism inspires a wave of research that unfolds along three main threads.

The first thread focuses on expressiveness and optimisation. Methods like AdaLoRA~\cite{Zhang2023AdaLoRAAB} and ReLoRA~\cite{Lialin2023ReLoRAHT} replace LoRA's static rank with dynamic allocation schedules to place capacity where it is most needed. Others enrich the update structure beyond a simple matrix product; for instance, BTT~\cite{Potapczynski2024SearchingFE} and CoMERA~\cite{Yang2024CoMERACA} introduce tensor factorisations while maintaining linear complexity. The training process itself is also refined. LoRA+~\cite{Hayou2024LoRAEL} advocates for distinct learning rates for the two factors, while LORO~\cite{LORO_iclr2025} and ScaleAdam~\cite{Zhang2024RiemannianPL} reframe the optimisation on Riemannian manifolds to align updates with the parameter space's intrinsic curvature.

A second thread pursues efficiency under tight resource constraints. QLoRA~\cite{Dettmers2023QLoRAEF} enables fine-tuning on a single consumer GPU by quantising the model backbone to 4-bits while keeping LoRA activations in higher precision. Complementary efforts further shrink memory footprints by compressing the adapters or sharing low-rank factors across layers and tasks, as seen in methods like LoQT and VeRA~\cite{Loeschcke2024LoQTLA,Kopiczko2023VeRAVR}.

The third thread involves the principled initialisation of weights. PiSSA~\cite{Meng2024PiSSAPS} and OLoRA~\cite{Bykakyz2024OLoRAOL} initialise the low-rank factors using singular vectors or orthogonal bases from the pre-trained weights, aligning the initial update with salient directions. DoRA~\cite{Liu2024DoRAWL} extends this by decomposing each weight into magnitude and direction, applying LoRA only to the directional component. Such targeted adaptation yields significant gains and highlights the value of structuring the update, not just constraining its rank.

Beyond LoRA-centric innovations, other structured PEFT methods explore different constraints. One class of methods, including HRA~\cite{Yuan2024BridgingTG} and HOFT~\cite{Arcas2025HOFTHO}, constructs complex orthogonal matrices from simple low-rank operations like Householder transformations. Another evolutionary line of structured methods seeks to balance expressive power, cost, and efficiency. This line evolves from the restrictive block-diagonal form of OFT~\cite{Qiu2023ControllingTD}, to the denser but more expensive butterfly structure of BOFT, and finally to the efficient group-and-shuffle structure of GSOFT~\cite{Gorbunov2024GroupAS}. Our method, SORA, also operates in this space. Inspired by the Cayley transform~\cite{Trockman2021OrthogonalizingCL}, SORA constructs orthogonal matrices using a first-order Taylor expansion of the matrix exponential, an approximation also explored in~\cite{Singla2021SkewOC}. As a future direction, SORA can be integrated with orthogonalisation techniques to enhance performance.

\section{Conclusion}
This paper identifies singular value entropy as the key mechanism behind DoRA's success leading to a unified weight conditioning framework for PEFT. Our framework yields two novel methods: \textit{(i)} \textbf{Pre-Diag}, which effectively calibrates pre-trained weights by applying diagonal matrix conditioning prior to the LoRA update, enhancing both training speed and performance; and \textit{(ii)} \textbf{SORA}, which introduces a parameter-efficient orthogonal rotation to perform a more powerful, norm-preserving transformation of the feature space, further improving model performance. Extensive experiments demonstrate that both methods achieve superior performance and computational efficiency over LoRA and DoRA. Ultimately, this research advocates for a paradigm shift in PEFT, moving from simple low-rank updates to more principled, structured optimizations of the weight space.
\section{Acknowledgments}
This research was supported by Guangdong S\&T Programme (No. 2024B0101010003). The computing resources were provided by Pengcheng Cloud Brain.

\bibliography{aaai2026}
\newpage
\appendix
\section*{\LARGE Appendix}
The appendices are structured as follows:
\begin{itemize}
    \item Appendix 1 offers the formal proof of Theorem 1.
    \item  Appendix 2 offers the formal proof of Theorem 2.
    \item Appendix 3 supplies additional details regarding the experimental setup.
\end{itemize}

\section{Appendix 1: Proof of Theorem 1}
\label{appendix:a}
\begin{proof}
For the update matrix $\Delta \mathbf{W}$ of LoRA, its set of normalized singular values is given by:
$$
\Sigma_{\text{LoRA}} = \{1, \underbrace{\alpha, \dots, \alpha}_{r-1}\}.
$$

For the update matrix $\Delta \mathbf{W}$ of DoRA, its set of normalized singular values is:
$$
\Sigma_{\text{DoRA}} = \{1, \underbrace{\beta, \dots, \beta}_{r-1}, \underbrace{\gamma, \dots, \gamma}_{s-r}\},
$$
where $s > r$.
Thus, we have
$$
\begin{aligned}
H_1 &= H(\sigma(\Delta \mathbf{W}_{\text{LoRA}})) \\
&= -\sum_{i=1}^{r} p_i \log(p_i) \\
&= -\left( p_{1,\max} \log(p_{1,\max}) + (r-1) p_{1,\alpha} \log(p_{1,\alpha}) \right) \\
&= -\left( \frac{1}{E_1} \log\left(\frac{1}{E_1}\right) + (r-1) \frac{\alpha^2}{E_1} \log\left(\frac{\alpha^2}{E_1}\right) \right),
\end{aligned}
$$

where $E_1 = 1^2 + (r-1)\alpha^2 = 1 + (r-1)\alpha^2$. The probability of the largest singular value $\sigma_1$ is $p_{1,\max}$, and the probability of the remaining singular values $\sigma_i$ (for $2 < i \le r$) is $p_{1,\alpha}$, with a total of $r-1$ such values.

Then, we have
$$
\begin{aligned}
H_2 &= H(\sigma(\Delta \mathbf{W}_{\text{DoRA}})) = -\sum_{i=1}^{s} p_i \log(p_i) \\
&= - p_{2,\max} \log(p_{2,\max})- (r-1) p_{2,\beta} \log(p_{2,\beta}) \\ & - (s-r) p_{2,\gamma} \log(p_{2,\gamma})  \\
&= - \frac{1}{E_2} \log\left(\frac{1}{E_2}\right) - (r-1) \frac{\beta^2}{E_2} \log\left(\frac{\beta^2}{E_2}\right)\\
&- (s-r) \frac{\gamma^2}{E_2} \log\left(\frac{\gamma^2}{E_2}\right),
\end{aligned}
$$
where $E_2 = 1 + \left( (r-1)\beta^2 + (s-r)\gamma^2 \right)$. The probability of the largest singular value $\sigma_1$ is $p_{2,\max}$; the probability corresponding to the singular value $\beta$ is $p_{2,\beta}$, with a total of $r-1$ such values; and the probability corresponding to the singular value $\gamma$ is $p_{2,\gamma}$, with a total of $s-r$ such values.

Since the sum of probabilities is 1, we have:
$$
\frac{1 + (r-1)\beta^2 + (s-r)\gamma^2}{E_2} = \frac{1 + (r-1)\alpha^2}{E_1} = 1.
$$

Thus:
$$
(1 + (r-1)\beta^2 + (s-r)\gamma^2) = \frac{E_2}{E_1} (1 + (r-1)\alpha^2).
$$

Furthermore, given our assumption that: $(r-1)\alpha^2 = (r-1)\beta^2 + (s-r)\gamma^2$ it follows that $E_1 = E_2$.

Next, we aim to prove that $H_2 > H_1$, i.e., to show that:
$$
\begin{aligned}
   - \left( (r-1)\frac{\beta^2}{E}\log\frac{\beta^2}{E} + (s-r)\frac{\gamma^2}{E}\log\frac{\gamma^2}{E} \right) \\> - \left( (r-1)\frac{\alpha^2}{E}\log\frac{\alpha^2}{E} \right) .
\end{aligned}
$$

Rearranging terms, we obtain:
$$
\begin{aligned}
&H_2-H_1 \\
&= (r-1)\alpha^2(\log\alpha^2 - \log E) - (r-1)\beta^2(\log\beta^2 - \log E) \\
& \quad - (s-r)\gamma^2(\log\gamma^2 - \log E) \\
&= (r-1)\alpha^2\log \alpha^2 - (r-1)\beta^2\log \beta^2 - (s-r)\gamma^2 \log \gamma^2 \\
& \quad + \left( (r-1)\beta^2 + (s-r)\gamma^2 - (r-1)\alpha^2 \right)\log E \\
&= (r-1)\alpha^2\log \alpha^2 - (r-1)\beta^2\log \beta^2 - (s-r)\gamma^2 \log \gamma^2 \\
&\stackrel{(\circ)}{=} \left( (r-1)\beta^2 + (s-r)\gamma^2 \right)\log\alpha^2 - (r-1)\beta^2\log\beta^2 \\
& \quad - (s-r)\gamma^2\log\gamma^2 \\
&\stackrel{(\star)}{=} (r-1)\beta^2 \log\left(\frac{\alpha^2}{\beta^2}\right) + (s-r)\gamma^2 \log\left(\frac{\alpha^2}{\gamma^2}\right) > 0,
\end{aligned}
$$
where $(\circ)$ follows from substituting the constraint $(r-1)\alpha^2 = (r-1)\beta^2 + (s-r)\gamma^2$ for $\alpha^2$ in the above expression; $(\star)$ follows from the constraint $(r-1)\alpha^2 = (r-1)\beta^2 + (s-r)\gamma^2$, from which we know that $\alpha^2 = \beta^2 + \frac{s-r}{r-1}\gamma^2$. Since $s > r$ and $\gamma \neq 0$, we have $\frac{s-r}{r-1}\gamma^2 > 0$. Therefore, $\alpha^2 > \beta^2 > \gamma^2$. This implies that $\frac{\alpha^2}{\beta^2} > 1$ and $\frac{\alpha^2}{\gamma^2} > 1$, so $\log\left(\frac{\alpha^2}{\beta^2}\right) > 0$ and $\log\left(\frac{\alpha^2}{\gamma^2}\right) > 0$.

Thus:
$$
H(\sigma(\Delta \mathbf{W}_{\text{DoRA}})) > H(\sigma(\Delta \mathbf{W}_{\text{LoRA}}))
$$

\end{proof}

\section{Appendix 2: Proof of Theorem 2}
\label{appendix:b}
\begin{proof}
Regarding the first formula:

\begin{equation}
\label{soc}
    \|\exp(\mathbf{J}) - \mathbf{S}_k(\mathbf{J})\|_2 \leq \frac{\|\mathbf{J}\|_2^k}{k!}\quad \text{where}\quad\mathbf{S}_k(\mathbf{J}) = \sum_{i=0}^{k-1}\frac{\mathbf{J}^i}{i!}.
\end{equation}

See the proof of [\cite{Singla2021SkewOC}, Theorem 3].

Next, we prove the second formula. First, we have:
$$
\mathbf{J} = \mathbf{D}\mathbf{C}^{\top} - \mathbf{C}\mathbf{D}^\top.
$$
Then we have:
$$
\begin{aligned}
\|\mathbf{J}\|_F &\stackrel{(\circ)}\le \|\mathbf{D}\mathbf{C}^\top\|_F + \|\mathbf{C}\mathbf{D}^\top\|_F\\
&\stackrel{(\star)}\le 2\|\mathbf{D}\|_F\|\mathbf{C}\|_F,
\end{aligned}
$$
where $(\circ)$ is due to the use of the triangle inequality; $(\star)$ is due to $\|\mathbf{A}\mathbf{B}\|_F \le \|\mathbf{A}\|_F\|\mathbf{B}\|_F$ for all $\mathbf{A} \in \mathbb{R}^{m \times n},\mathbf{B} \in \mathbb{R}^{n \times m}$.

Let $s_p = \frac{\epsilon}{2\|\mathbf{D}\|_F\|\mathbf{C}\|_F + \epsilon}$ and $\mathbf{P} = s_p\mathbf{J}$. From formula \ref{soc}, we obtain:
$$
\begin{aligned}
\|\exp(\mathbf{P}) - \mathbf{S}_1(\mathbf{P})\|_2 &\leq \|\mathbf{P}\|_2\\
&\le \|\mathbf{P}\|_F \\
&= \|s_p\mathbf{J}\|_F \\
&\le s_p \|\mathbf{J}\|_F \\
&\le \epsilon.
\end{aligned}
$$
\end{proof}

\section{Appendix 3: Experimental Details}
\label{appendix:e}

\subsection{GLUE Benchmark}
We fine-tune the pre-trained DeBERTa-Base model\cite{He2020DeBERTaDB} on the GLUE benchmark using the Hugging Face implementation\footnote{\url{https://huggingface.co/transformers/model_doc/roberta.html}}\footnote{\url{https://huggingface.co/datasets/nyu-mll/glue}}. For all tasks except QQP, we employ a batch size of 32, while QQP uses a larger batch size of 128 due to its dataset characteristics. The model is trained uniformly for 5 epochs across all tasks with a maximum sequence length of 128. For each task, we perform a grid search over learning rates, and for each learning rate, we run three independent random experiments. For most methods, the learning rate range is $\{1\text{e-}3, 8\text{e-}4, 4\text{e-}4\}$ and the weight decay is set to 0.01. 


\subsection{Commonsense Benchmark}
We fine-tune the pre-trained LLaMA3-8B model \cite{Dubey2024TheL3} using the LLM-Adapters codebase\footnote{\url{https://github.com/AGI-Edgerunners/LLM-Adapters}} with evaluation via standardized lm-evaluation-harness\footnote{\url{https://github.com/EleutherAI/lm-evaluation-harness}} on the Commonsense benchmark with the Hugging Face 
implementation\footnote{\url{https://huggingface.co/datasets/fw407/Commonsense-15K}}. The fine-tuning process employs consistent hyperparameters across the AdamW optimizer. Specifically, we train for 5 epochs with a total of 588 training steps, including 20 warm-up steps. The batch size is set to 128, and the maximum sequence length is 512. We use a learning rate of $4\text{e-}4$ and optimizer parameters $\beta_1 = 0.9$ and $\beta_2 = 0.999$. We evaluate the fine-tuned model on eight commonsense reasoning tasks: BoolQ~\cite{Clark2019BoolQET}, PIQA~\cite{Bisk2019PIQARA}, SIQA~\cite{Bisk2019PIQARA}, HellaSwag~\cite{zellers2019hellaswag}, WinoGrande~\cite{Sakaguchi2019WinoGrande}, ARC-e and ARC-c~\cite{Clark2018ThinkYH}, and OpenbookQA~\cite{Mihaylov2018CanAS}. 

\subsection{Mathematical Benchmark}
We fine-tune the pre-trained Gemma-7B model \cite{gemma_2024} using the LLM-Adapters codebase\footnote{\url{https://github.com/AGI-Edgerunners/LLM-Adapters}} with evaluation via standardized lm-evaluation-harness\footnote{\url{https://github.com/EleutherAI/lm-evaluation-harness}} on the subset of the Mathematical benchmark with the Hugging Face 
implementation\footnote{\url{https://huggingface.co/datasets/meta-math/MetaMathQA}}. We fine-tune the model using the AdamW optimizer. The training runs for 2 epochs (192 total steps, with 8 warm-up steps), using a batch size of 128 and a maximum sequence length of 512. We perform a hyperparameter search over the learning rate from the set $\{2\text{e-}4, 4\text{e-}4, 6\text{e-}4\}$, while keeping the optimizer parameters fixed at $\beta_1 = 0.9$ and $\beta_2 = 0.999$. The model is evaluated on six mathematical reasoning benchmarks: GSM8K~\cite{cobbe2021training}, MultiArith~\cite{roy2016solving}, AQuA~\cite{ling2017program}, SVAMP~\cite{patel2021nlp}, AddSub~\cite{hosseini2014learning}, and SingleEq~\cite{koncel2015parsing}.
\section{Limitations}
This work is subject to two main limitations. First, our analysis focuses on establishing the prerequisite correlation between SVD entropy and performance (Figure.\ref{Fig:G2}, Table \ref{table-glue},\ref{table-commonsense},\ref{table-math}), rather than directly proving causality. We acknowledge that designing a valid spectral regularizer to formally test this causal relationship is a non-trivial research challenge. However, our findings establish a crucial foundation that enables this direction for future work. Second, our empirical validation is currently confined to language models. The applicability and performance of our SORA method on fine-tuning vision models have not yet been thoroughly investigated.
\end{document}